\documentclass{article}

\usepackage{amsmath,amsfonts,bm}

\def\Figref#1{Figure~\ref{#1}}

\def\Secref#1{Section~\ref{#1}}

\def\eqref#1{equation~\ref{#1}}

\def\1{\bm{1}}

\DeclareMathAlphabet{\mathsfit}{\encodingdefault}{\sfdefault}{m}{sl}
\SetMathAlphabet{\mathsfit}{bold}{\encodingdefault}{\sfdefault}{bx}{n}

\usepackage[sort,authoryear,round]{natbib}

\usepackage{siunitx}

\usepackage{fullpage}

\usepackage[utf8]{inputenc} 
\usepackage[T1]{fontenc}    
\usepackage{xcolor}
\usepackage{hyperref}       
\definecolor{myblue}{rgb}{0,0.2,0.8}
\hypersetup{ %
pdftitle={},
pdfauthor={},
pdfsubject={},
pdfkeywords={},
pdfborder=0 0 0,
pdfpagemode=UseNone,
colorlinks=true,
linkcolor=myblue,
citecolor=myblue,
filecolor=myblue,
urlcolor=myblue,
pdfview=FitH}
\usepackage{authblk}
\usepackage{url}            
\usepackage{booktabs}       
\usepackage{amsfonts}       
\usepackage{amsthm}
\usepackage{nicefrac}       
\usepackage{microtype}      
\usepackage{amsmath} 

\usepackage{graphicx}
\usepackage{subcaption}
\usepackage{fancyvrb}

\usepackage{xspace}
\usepackage[varqu,varl,var0,scaled=0.97]{inconsolata}
\usepackage{overpic}
\usepackage{wrapfig}
\usepackage{tcolorbox}
\usepackage{enumitem}
\usepackage{threeparttable}

\usepackage{algorithm}
\usepackage{algpseudocode}

\usepackage{listings}
\usepackage{color}
\usepackage{lipsum}

\definecolor{dkgreen}{rgb}{0,0.6,0}
\definecolor{gray}{rgb}{0.5,0.5,0.5}
\definecolor{mauve}{rgb}{0.58,0,0.82}

\newcommand{\eg}{\emph{e.g.},\xspace}
\newcommand{\ie}{\emph{i.e.},\xspace}

\newcommand{\tabref}[1]{Table~\ref{#1}}

\newcommand\circled[1]{\textcircled{\small{#1}}}

\usepackage[colorinlistoftodos,textsize=tiny,textwidth=15mm]{todonotes}

\title{
\bf{Geometric Data Augmentations to Mitigate Distribution Shifts in Pollen Classification from Microscopic Images
}}

\author[1,2]{Nam Cao}
\author[1,3]{Olga Saukh}

\affil[1]{Institute of Technical Informatics, Graz University of Technology, Austria}
\affil[2]{University of Technology and Education, University of Da Nang, Vietnam}
\affil[3]{Complexity Science Hub Vienna, Austria}

\affil[ ]{}
\affil[ ]{\texttt{
\{cao.nam, saukh\}@tugraz.at}
}

\begin{document}
\date{}
\maketitle

\begin{abstract}
Distribution shifts are characterized by differences between the training and test data distributions. They can significantly reduce the accuracy of machine learning models deployed in real-world scenarios. This paper explores the distribution shift problem when classifying pollen grains from microscopic images collected in the wild with a low-cost camera sensor. We leverage the domain knowledge that geometric features are highly important for accurate pollen identification and introduce two novel geometric image augmentation techniques to significantly narrow the accuracy gap between the model performance on the train and test datasets. In particular, we show that Tenengrad and ImageToSketch filters are highly effective to balance the shape and texture information while leaving out unimportant details that may confuse the model. Extensive evaluations on various model architectures demonstrate a consistent improvement of the model generalization to field data of up to $14\%$ achieved by the geometric augmentation techniques when compared to a wide range of standard image augmentations. The approach is validated through an ablation study using pollen hydration tests to recover the shape of dry pollen grains. The proposed geometric augmentations also receive the highest scores according to the affinity and diversity measures from the literature.
\end{abstract}


\section{Introduction}

Distribution shifts pose significant challenges in various data analysis tasks, including pollen sensing. These shifts occur when the distribution of data in the training set differs from the distribution in the target environment, leading to a reduced model performance. The problem has received significant attention in the recent scientific literature~\citep{Hendrycks2020, koh2021wilds, Yao2022, taori2020measuring} since it often prevents using deep learning models in many practical use-cases ~\citep{Jean2016, zech2018, lazaridou2021mind}.

The four main groups of approaches used to tackle the distribution shift problem are pre-training, data augmentation~\citep{liu2022empirical}, designing invariant architectures~\citep{Biscione2021}, and domain adaptation methods~\citep{li2019target, 8610281}. In the era of data, the dataset design and data filtering is also in focus of the recent community effort providing a testbed for dataset-related experiments and a systematic way to evaluate the model robustness on distribution shifts ~\citep{gadre2023datacomp}.

On the application side, numerous studies have been investigating the data distribution shift problem in real-world datasets, with a particular focus on the diversity across demographic groups, users, hospitals, camera locations, countries, time periods, and molecular scaffolds, as surveyed in~\citep{koh2021wilds}. In the context of pollen classification from microscopic images, the existing studies~\citep{gonccalves2016feature, battiato2020pollen13k,  Sevillano2020, Chen2022} achieve high accuracy up to 97\% on test data from the same distribution. 
However, the datasets used in these studies are limited in diversity and scale, making it challenging to conduct robustness assessments.

\paragraph{Contributions.}
In this paper, we address the distribution shift problem in the domain of automatic pollen identification from microscopic images by introducing novel geometric image augmentation techniques. We use a pollen dataset gathered by a fully automated sampling station operating in two modes: (i) processing curated pollen grains, and (ii) running in the wild. The dataset exhibits a noticeable distribution shift between both operation modes. We make the following contributions:

\begin{itemize}
\item Motivated by the expert knowledge regarding the importance of geometry and texture in pollen identification, we propose a novel geometric image augmentation approach, which effectively balances the shape and the texture information in pollen data. We show that the proposed method consistently improves pollen classification performance from microscopic images across multiple CNN architectures.

\item We conduct an ablation study to assess the model's ability to generalize to the correct pollen grain data by making use of the pollen hydration process, which is known to restore geometry and internal structures of dry pollen. We observe that geometric augmentations are consistently more useful as hydration advances.

\item Finally, we show that the proposed techniques improve the affinity and diversity measures from the literature, proposed to quantify the impact of data augmentation on the model generalization ability. 
\end{itemize}

The next section summarizes related work on the topic. \Secref{sec:Datasets} introduces pollen datasets and discusses the distribution shift problem when it comes to training a robust pollen classification model. \Secref{sec:geometric_augmentations} presents the proposed geometric image augmentation methods: Tenengrad and ImageToSketch. We then extensively evaluate the performance of the proposed techniques in isolation and in combination in \Secref{sec:experiments}. \Secref{sec:conclusion} summarizes our findings, lists limitations and outlines our future research directions.

\section{Related work}
\label{sec:related}
Data distribution shifts hamper the use of modern deep learning models in real-world use-cases, despite their superior performance over conventional baselines and even human experts. We first describe the natural distribution shift problem and then survey the state-of-the-art methods from the literature used to tackle the problem.

\subsection{Natural distribution shifts}
Building a robust machine learning model is a significant challenge due to the presence of distribution shifts \citep{koh2021wilds, Yao2022, taori2020measuring, stacke2019closer, Stacke2021}.
Conventional datasets, such as ImageNet~\citep{Deng09ImageNet} and CIFAR10~\citep{cifar100}, have commonly been used to design and evaluate deep neural networks, which led to significant improvement of deep learning methods over the years. Model evaluation on real-world data, however, often presents an additional challenge due to the presence of various sources of distribution shift, including lighting conditions, camera orientations, sensor mounting, different viewpoints and styles, temporal~\citep{Yao2022} and domain shifts~\citep{stacke2019closer}.

In the context of histopathology image classification, studies have explored the internal representation learned by convolutional neural networks (CNN) and introduced measures to quantify model-specific domain shifts~\citep{stacke2019closer, Stacke2021}. Despite the ongoing research efforts, there is currently no universal algorithm to handle distribution shifts, and the choice of measures often depends on the dataset-specific characteristics. 

Complementing the existing literature on the data distribution shifts~\citep{Hendrycks2020, taori2020measuring}, we chose a compact dataset to study the effectiveness of data augmentation methods. This choice is rooted in several key factors that enhance the robustness of our analysis. Unlike datasets with a wide range of context and environmental conditions, such as ImageNet, a pollen dataset offers a controlled environment to test methods that minimize the natural distribution shift. By carefully adjusting the lighting setting and focusing on individual grains without complex contextual elements or occlusions, we can ensure a clean capturing environment. The predominant distribution shifts within the gathered dataset primarily arise from biodiversity and varying hydration levels across input pollen grains, offering a unique setting to study data augmentation effects to tackle natural distribution shifts. Moreover, conducting ablation studies is straightforward, allowing us to isolate and analyze the impact of individual augmentation techniques on model performance. A detailed overview of the dataset is provided in~\Secref{sec:Datasets}.

\begin{figure}[t]
  \centering
  \includegraphics[width=0.65\linewidth]{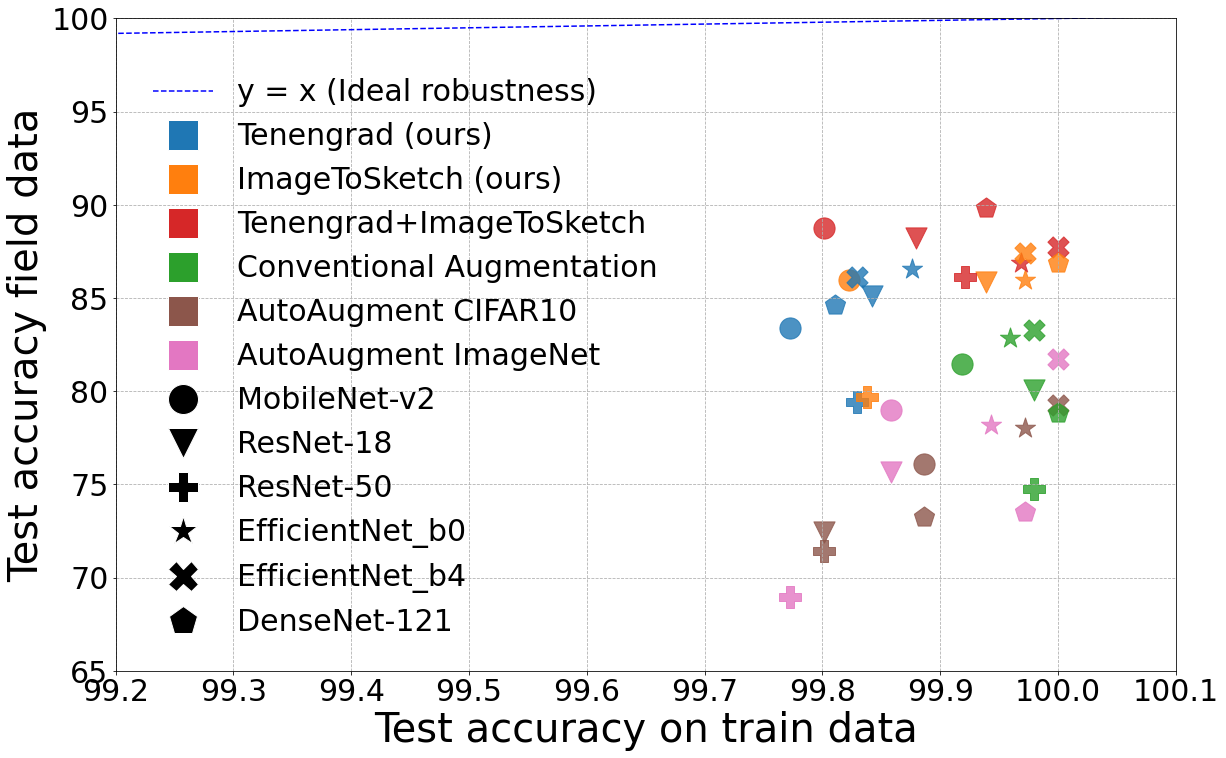}
  \caption{Distribution shifts cause accuracy drop. The models feature different CNN architectures (same marker shape), while different augmentation methods (same color) are used during training. All models are evaluated on the test library data and on the field data. Geometric augmentations (blue, orange and red) outperform other methods.}
  \label{fig:distribution_shift}
\end{figure}

\subsection{Mitigating distribution shifts}
\label{sec:fighting_distribution_shifts}
There are four key approaches commonly used to tackle the distribution shift problem summarized below.

\paragraph{Pre-training.} 
Model pre-training on a large source dataset with abundant labeled data has shown to outperform training from scratch when fine-tuned on a downstream task, especially if the downsteam dataset contains only very few samples~\citep{entezari2023role}. The pre-trained model learns good features and representations from the upstream dataset~\citep{NEURIPS2020_0607f4c7}. Once pre-training is complete, the model is fine-tuned on a smaller dataset from the target domain, where the available representations appear useful. This two-step approach is widely used in practice and is particularly effective when the source and the target domains share some similarities~\citep{entezari2023role}. In this research, we employ models pre-trained on ImageNet \citep{Deng09ImageNet}.

\paragraph{Data Augmentation.}
Data augmentation is a technique used to artificially increase the diversity of the training data by applying various transformations to the original samples~\citep{shorten_survey_2019}.
By introducing these variations, the model becomes more robust to changes in the data distribution during testing. Data augmentation is particularly useful when the target domain has limited labeled data or when the data in the target domain is scarce and different from the source domain. Various well-known image augmentation methods have been proposed ranging from conventional augmentations using rotation, affine and color transformations, to more advanced methods such as AutoAugment~\citep{Cubuk2018}, MixUp~\citep{zhang2018mixup}, and CutMix~\citep{yun2020cutmix}. 
In this work, we show that these methods do not appear effective in the context of pollen classification, because pollen morphology such as shape, texture, and aperture are precious details, and important to correctly classify different pollen taxa. Our experiments also show that the augmentation methods that contain the image deformation and crop action such as AutoAugment, Mixup, and CutMix do not work with pollen image data. In the remaining parts of this paper, we only compare the AutoAugment method and Conventional augmentations in our experiments. We propose in \Secref{sec:geometric_augmentations} a new augmentation method that emphasizes the geometric features to mitigate the distribution shift and increase model robustness.

\paragraph{Designing Invariant Architectures.}
An invariant architecture is designed to be robust to specific transformations of the input data. For example, a CNN with spatial pooling layers can classify images of objects regardless of their location in the scene, \ie CNN implements the desired invariance to translation (for the sake if this example we omit the criticism of CNN's translation invariance~\citep{Biscione2021}). 
Invariant architectures for a specific transformation are often manually designed and are known to outperform data augmentations~\citep{sabour2017dynamic}.

\paragraph{Data Adaptation Methods.}
Data adaptation methods aim to bridge the gap between the source and target domains by minimizing the discrepancy between their distributions. Various techniques fall under this category, such as domain adaptation~\citep{8610281, li2019target} and domain generalization~\citep{Hou_2022_CVPR, rosenfeld2022domainadjusted}. Domain adaptation attempts to align the feature distributions of the source and target domains by learning domain-invariant features. Domain generalization, on the other hand, deals with adapting a model to multiple source domains, so it can generalize well to a new, unseen target domain. These methods are mostly based on Generative Adversarial Networks (GANs), which are out-of-scope of this paper.

\section{Pollen Sensing and Pollen Datasets}
\label{sec:Datasets}
Our study of distribution shifts in microscopic pollen images is motivated by the following observations: (i) Although existing pollen classifiers achieve high accuracy of up to 97\,\%~\citep{Sevillano2020, Chen2022}, the existing datasets used in these studies are frequently limited in the number of samples and diversity of grain views with respect to both depth of view and grain orientation, making it challenging for a model to learn pollen morphology. (ii) We make use of the publicly available Graz pollen dataset~\citep{Cao2020PollenDataset} of pollen taxa gathered with a fully automated low-cost sensing device. We use the same device to run hydration tests to highlight the mechanism that allows us achieving significant performance gains by our data augmentation techniques.

Different technologies have been recently developed for automated pollen sensing and identification, mostly divided into image-based and non-image-based methods. Non-image-based methods use alternative techniques for sensing pollen characteristics, for example, fluorescence \citep{MITSUMOTO200990}, Fourier-Transform infrared \citep{dellanna_pollen_2009}, and Raman spectroscopy \citep{ivleva_characterization_2005}, etc. The two main image-based methods are light microscopic imaging and holography imaging~\citep{Sauvageat2019}. We focus on pollen sensing methods based on light camera imaging and grain classification using machine learning methods. The main advantage of this approach is that human experts can visually inspect the image and verify the performance of the automated classifier. Data collection by such pollen sensing stations is performed using multiple steps: sample collection, sample pre-processing, sample scanning, layer-wise focusing and image capturing. The sample pre-processing step sometimes includes the staining step to highlight structural details of pollen grains~\citep{astolfi2020pollen73s}. Staining is a good support for human vision, as it helps to increase the contrast and highlight details. For deep learning algorithms, it is one of the sources of distribution shift, since the staining color is not uniform across different samples and experiments. The image acquisition is done manually or by a commercial automatic device~\citep{Sevillano2020}. This dataset generation pipeline results in a uniformly sampled high-quality dataset but is labor-intensive to create and difficult to reproduce. There are several publicly available datasets of microscopic images of pollen, \eg Pollen23E~\citep{gonccalves2016feature}, Pollen13K~\citep{battiato2020pollen13k},
Pollen73S~\citep{astolfi2020pollen73s},
New Zealand Pollen dataset~\citep{Sevillano2020}, and Graz pollen dataset~\citep{Cao2020PollenDataset}. 

The Graz pollen dataset used in this work was acquired automatically with an inexpensive station~\citep{Cao2020,Cao2019} and without sample pre-processing, staining and scanning. The dataset comprises approximately 16'000 samples distributed among five taxa: Alnus, Betula, Carpinus, Corylus, and Dactylis.
It includes both the data obtained using manually curated pollen grains (referred to as the library data) and the field data. The library data was collected from specific trees, whereas the field data was labeled by an expert.
The details on the number of samples per taxa can be found in~\tabref{tab:TrainAndFieldData}. In contrast to the library data, which is of high quality, field data was collected in the wild by automatically running the station for 10 weeks. The Graz pollen dataset is thus well-suited to study distribution shifts that occur when training a model on the library images and testing it on the field data collected with the same station. We measured an average classification accuracy drop of around 20--25\,\% across various model architectures, including Mobilenet-v2~\citep{sandler2018mobilenetv2}, ResNet-18~\citep{he2016deep}, ResNet-50~\citep{he2016deep}, EfficientNet-b0~\citep{Tan2019}, EfficientNet-b4~\citep{Tan2019} and DenseNet-121 ~\citep{huang2017densenet}. 
\Figref{fig:distribution_shift} plots test set accuracy of different models on the library data against models' performance on the field data. The $y = x$ line is the ideal robustness line~\citep{pmlr-v97-recht19a}: the closer the model performance is to $y = x$ the higher is its effective robustness.

Upon analyzing the field data, we identified two significant factors impacting the performance of our model. Firstly, we observed the presence of samples of the same taxa in the library but in a slightly different view on the aperture, the outer shape (exine), orientation of grains, and color. We attribute these differences to the local biodiversity in the field. Secondly, variations in lighting conditions arise due to differences in the glycerine levels. The automatic pollen sensor used to collect the Graz pollen dataset makes use of glycerine for better image capture and for hydration of pollen to recover its morphology before being processed. The importance of pollen hydration will be explained in \Secref{sec:pollen_hydration}.  Furthermore, the limited depth of view in microscopic images of a few dozen micrometers due to the use of low-cost hardware components results in distinct appearances of captured grains at different depths. Together, these factors cause distribution shift of the field data, leading to a drop of model performance. 
Irrelevant particles contained in the recorded data were removed from the dataset.
In the following section, we discuss the geometric data augmentation techniques we use to increase models' effective robustness on pollen field data.

\begin{figure*}[t]
\centering
    \begin{subfigure}{0.95\textwidth}
        \includegraphics[width = \textwidth]{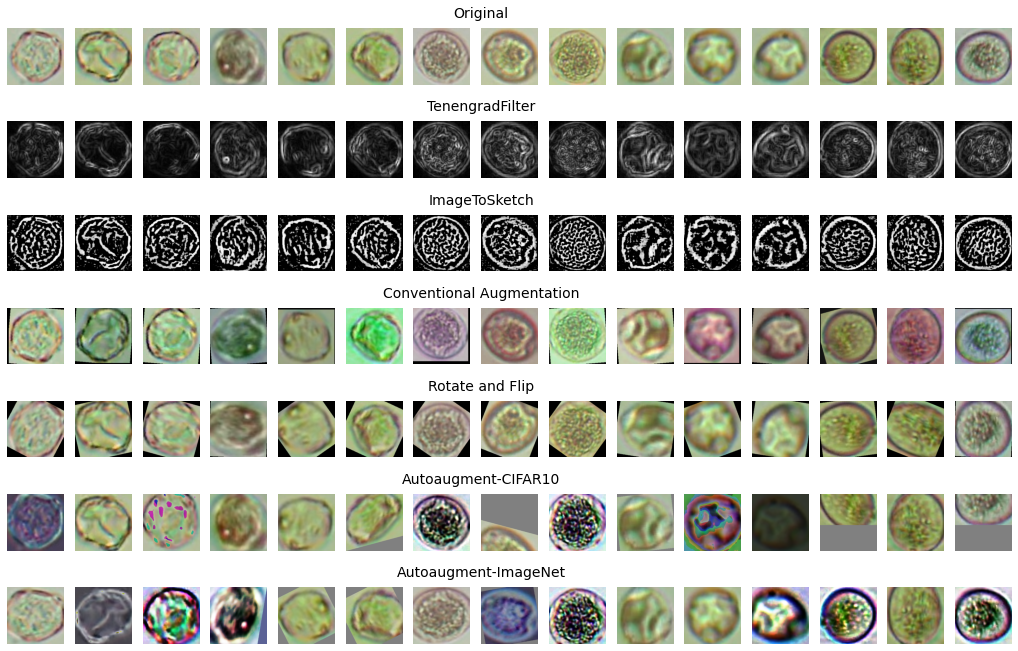}
    \end{subfigure}	
   	
    \caption{Sample images of applying various data augmentation methods used throughout this work. Tenengrad and ImageToSketch are novel augmentation techniques proposed in this work and used to promote shape-specific features important for accurate pollen identification.}
    \label{fig:augmentation_samples}
\end{figure*}

\begin{table}[]
\centering
\caption{\#samples in the library (train) and field (test) datasets.} 
\label{tab:TrainAndFieldData}

\begin{tabular}{lcccc}
\toprule
\textbf{} & \multicolumn{2}{c}{\textbf{Library data (train)}}               & \multicolumn{2}{c}{\textbf{Field data (test)}}                     \\ 
          & \multicolumn{1}{c}{\#samples} & \multicolumn{1}{c}{\%} & \multicolumn{1}{c}{\#samples} & \multicolumn{1}{c}{\%} \\ 
\midrule
Alnus          & \num{1804} & \num{25.74} & \num{1191}  & \num{7.22}  \\
Betula         & \num{517}  & \num{7.38}  & \num{4113}  & \num{24.94} \\
Carpinus       & \num{1314} & \num{18.75} & \num{3979}  & \num{24.13} \\
Corylus        & \num{1652} & \num{23.57} & \num{3554}  & \num{21.55} \\
Dactylis       & \num{1722} & \num{24.57} & \num{3654}  & \num{22.16} \\ \midrule
\textbf{Total} & \num{7009} &   100    & \num{16494} &  100 \\
\bottomrule
\end{tabular}
\end{table}

\section{Geometry-based Data Augmentation}
\label{sec:geometric_augmentations}
In this section, we first give reasons why including shape information helps to boost relevant features during model training. We then introduce our two geometry-based data augmentation techniques.

\subsection{Texture or shape}
Understanding and incorporating shape information is crucial for training robust deep learning models. As highlighted in \citep{Cai2019}, the analysis of CNNs trained on large-scale datasets reveals a bias towards texture features, indicating a tendency to prioritize texture cues when making predictions. 
According to \citet{Islam2021}, lower-level features extracted from early layers of the network, predominantly capture texture cues. Unfortunately, these layers exhibit a lower prevalence for shape-related features. As the network progresses to higher layers, a gradual incorporation of shape information becomes apparent, although texture features still dominate. This bias leads to limited accuracy and robustness of CNNs on various tasks.
The findings by \citet{Cai2019} and \citet{Islam2021} collectively highlight the significance of balancing texture and shape information during model training. By prioritizing shape information, deep learning models can effectively capture the intricacies of objects and scenes, leading to more accurate and robust predictions. Incorporating shape information in training helps to mitigate the limitations imposed by the texture bias. Based on the above evidence we propose an image augmentation method that promotes the shape information during training. The method is described in the following sections.

\subsection{Tenengrad Filter}
Tenengrad is a gradient magnitude metric first described in \citep{tenenbaum1970accommodation} and later explored in \citep{schlag1983implementation} and \citep{krotkov1988focusing}. Tenengrad combines magnitude information from two Sobel kernels~\citep{Sobel1968}. Sobel operator is a discrete differentiation operator and calculates the first-order image derivatives by convolving the image with a small kernel in the $x$ and $y$ directions. 
The Scharr kernels introduced in \citep{Scharr2000} are slightly different from the Sobel kernels and are optimized to provide more accurate gradient approximations, especially in the case of diagonal edges. Therefore, the Scharr filter captures more details and provides better edge detection results.

When applying data augmentation, grain rotations in the input image introduce geometric changes that affect the gradient calculation. The Scharr operator is more rotation-invariant compared to the Sobel operator due to its kernel design. Consequently, our approach draws inspiration from the Tenengrad filter but employs the Scharr operator with two kernels instead of the Sobel operator:

\vspace{6pt} 
\begin{minipage}[c]{.4\linewidth}
\[
\textbf{S}_x=
\begin{bmatrix}
 -3 & 0 & 3 \\
 -10 & 0 & 10 \\
 -3 & 0 & 3 
\end{bmatrix}
\]
\end{minipage}
\begin{minipage}[c]{.1\linewidth}
and 
\end{minipage}
\begin{minipage}[c]{.4\linewidth}
\[
\textbf{S}_y=
\begin{bmatrix}
 -3& -10 & -3 \\
 0 & 0 & 0 \\
 3 & 10 & 3 
\end{bmatrix}.
\]
\end{minipage}
\vspace{6pt} 

Assuming that an image to be operated is $\textbf{I}$, horizontal and vertical changes are computed by convolving $\textbf{I}$ with kernels $\textbf{G}_x$ and $\textbf{G}_y$ respectively: 

\begin{minipage}[c]{.4\linewidth}
\[
\textbf{G}_x = \textbf{S}_x \ast \textbf{I}
\]
\end{minipage}
and 
\begin{minipage}[c]{.4\linewidth}
\[
\textbf{G}_y = \textbf{S}_y \ast \textbf{I}.
\]
\end{minipage}
\vspace{6pt} 

At each point of the image $\textbf{I}$, we calculate an approximation of the gradient at that point by combining both results above
$$\textbf{G}=\sqrt{\textbf{G}_{x}^{2}+\textbf{G}_{y}^{2}}.$$
Image $\textbf{G}$ is then normalized to ensure that the array values are between 0 and 255. Line 2 in \Figref{fig:augmentation_samples} shows the result of applying the Tenengrad filter to sample input images.

\subsection{ImageToSketch Filter}
In this section, we introduce another geometric filter that also emphasizes the edges and shape details in the input image. This filter aims to destroy colors and background information, while highlighting the most contrasting details in the input. For an input gray image, we first apply Gaussian smoothing and then create a sketch effect by dividing the gray image by the smoothed image. The result is a sketched image with enhanced edges and a stylized appearance. The histogram equalization is then applied to enhance the contrast. This step is crucial for low-contrast images such as microscopic images of pollen grains. The final steps are to improve morphological opening, and include dilation and erosion to remove noise, eliminate small objects and smooth the edges. Algorithm \ref{alg:ImgToSketch} presents the steps of the ImageToSketch filter. The $\mathbf{kernel\_size}$ can be chosen based on input image size. Due to small image sizes in our experiments, the kernel sizes for $\mathbf{kernel\_size1}$ and $\mathbf{kernel\_size2}$ are set to $1$ and $2$, respectively. Sample images obtained by applying the filter are on line 3 of \Figref{fig:augmentation_samples}. Our source code\footnote{https://github.com/caonam-tugraz/geometric\_data\_augmentation} and the dataset\footnote{https://zenodo.org/records/10067085} used in the paper are available online.

\begin{algorithm}[h]
  \caption{ImageToSketch filter.}\label{alg:ImgToSketch}
  \begin{algorithmic}[1]
    \Require $\mathbf{img\_gray}$: Input image to be processed
    \Ensure $\mathbf{sketched\_img}$: Sketched image
    \State $\mathbf{smoothed\_img} \leftarrow \texttt{GaussianBlur}(\mathbf{img\_gray}, (21, 21))$
    \State $\mathbf{sketched\_img} \leftarrow \mathbf{img\_gray} / \mathbf{smoothed\_img}$
    \State $\mathbf{equalized} \leftarrow \texttt{equalizeHist}(\mathbf{sketched\_img})$
    \State Let $\mathbf{kernel1}$ be a ones-matrix with $\mathbf{kernel\_size1}$
    \State $\mathbf{morph} \leftarrow \texttt{morphologyEx}(\mathbf{equalized},\texttt{MORPH\_OPEN}, \mathbf{kernel1})$
    \State Let $\mathbf{kernel2}$ be a ones-matrix with $\mathbf{kernel\_size2}$
    \State $\mathbf{sketched\_img} \leftarrow \texttt{dilate}(\mathbf{morph}, \mathbf{kernel2})$
  \end{algorithmic}
\end{algorithm}

As next, we show the effectiveness of the presented filters when used as data augmentation to fine-tune models on the library data.

\section{Experiments}
\label{sec:experiments}
We run three experiments that prove efficiency of the proposed geometric filters. Our evaluation shows (1) consistently and significantly improved test set accuracy on 6 state-of-the-art deep model architectures, (2) consistent improvements with higher pollen hydration levels used to restore grain geometry, and (3) increased affinity and diversity measures suggested in the literature to assess effectiveness of data augmentation techniques. All images from the Graz pollen dataset are resized to 224x224 pixels in our experiments.

\subsection{Different CNN architectures, model sizes, and image augmentation methods}
\label{sec:train_on_different_CNN}
Our primary objective is to examine the influence of geometric image augmentation on the performance of pollen classification using state-of-the-art network architecture. Table \ref{tab:CNN_architectures} shows six CNN architectures with varying number of parameters ranging from 3.4 million (MobileNet-v2) to 23.9 million (ResNet-50), which are utilized in our experiments in \Secref{sec:train_on_different_CNN}. 

\begin{table}[t]
\centering
\caption{CNN architectures and number of parameters.}
\label{tab:CNN_architectures}
\begin{tabular}{lc}
\toprule
\textbf{CNN architecture} & \textbf{\#parameters [million]} \\ 
\midrule
MobileNet-v2 \citep{sandler2018mobilenetv2}        & 3.4                       \\
EfficientNet-b0 \citep{Tan2019}      & 5.3                       \\
DenseNet-121 \citep{huang2017densenet}          & 7.2  \\
ResNet-18 \citep{he2016deep}            & 11.2                      \\
EfficientNet-b4 \citep{Tan2019}      & 19.3                      \\
ResNet-50 \citep{he2016deep}             & 23.9                      \\
\bottomrule
\end{tabular}
\end{table}

All models are initially pre-trained on ImageNet~\citep{Deng09ImageNet} and then fine-tuned on the pollen library dataset for 40 epochs until convergence at around epoch 30-35. More details of training parameters can be found in Appendix~\ref{app:traning_parameters}.
Five distinct image augmentation techniques are applied during fine-tuning: Tenengrad and ImageToSketch augmentations, Conventional augmentation, AutoAugment-CIFAR10, and AutoAugment-ImageNet~\cite{Cubuk2018}. 
The Conventional augmentation is a sequence of image transformations used in \cite{krizhevsky2012imagenet} for ImageNet classification: 

\begin{Verbatim}[fontsize=\small]
Conventional = Resize + RandomRotate(10°) + RandomShift(x=0.05, y=0.05) + ColorJitter(brightness=0.2, 
contrast=0.3, saturation=0.3, hue=0.3) + Normalize.
\end{Verbatim}

\noindent Tenengrad augmentation comprises the following steps: 

\begin{Verbatim}[fontsize=\small]
Tenengrad = Resize + CenterCrop + HorizontalFlip(p) + RandomRotate(10°) + Tenengrad(p) + RandomShift(p, 
percent) + ColorJitter(brightness=0.2, contrast=0.3, saturation=0.3, hue=0.3) + Normalize.
\end{Verbatim}

\noindent ImageToSketch is a sequence of image transformations: 
\begin{Verbatim}[fontsize=\small]
ImageToSketch = Resize + CenterCrop + HorizontalFlip(p) + RandomRotate(10°) + ImageToSketch(p) + 
RandomShift(p, percent) + ColorJitter(brightness=0.2, contrast=0.3, saturation=0.3, hue=0.3) + Normalize.
\end{Verbatim}

In order to exploit the joint ability of the two latter methods, we combine the Tenengrad and ImageToSketch augmentations into a single sequence named \texttt{Tenengrad+ImageToSketch}, the probability of applying one of the augmentation methods is set to \texttt{p} = 0.5, this means when the \texttt{Tenengrad+ImageToSketch} filter is activated, the two filters Tenengrad and ImageToSketch are applied in an alternating manner.
We control the probability \texttt{p} of applying a step to create more diverse augmentations, in our case \texttt{p} is set to 0.4 for all experiments except the experiments for affinity and diversity in \Secref{sec:diversity_affinity}.

AutoAugment-ImageNet is a set of 24 best sub-policies of augmentation on ImageNet, the sub-policy is randomly chosen at run-time. Every sub-policy is constructed by combining various transformations, including \texttt{Posterize}, \texttt{Rotate}, \texttt{Solarize}, \texttt{Equalize}, \texttt{Invert}, \texttt{Color}, \texttt{Contrast}, \texttt{Autocontrast}, \texttt{Sharpness}, and \texttt{ShearX}. Similarly, AutoAugment-CIFAR10 is a set of 25 sub-policies that work best on the CIFAR10 dataset. Further information on AutoAugment can be found on GitHub~\citep{githubrepo_autoaugment}.

\begin{figure*}[t]
     \centering`
     \begin{subfigure}{0.325\linewidth}
        \includegraphics[width = \linewidth]{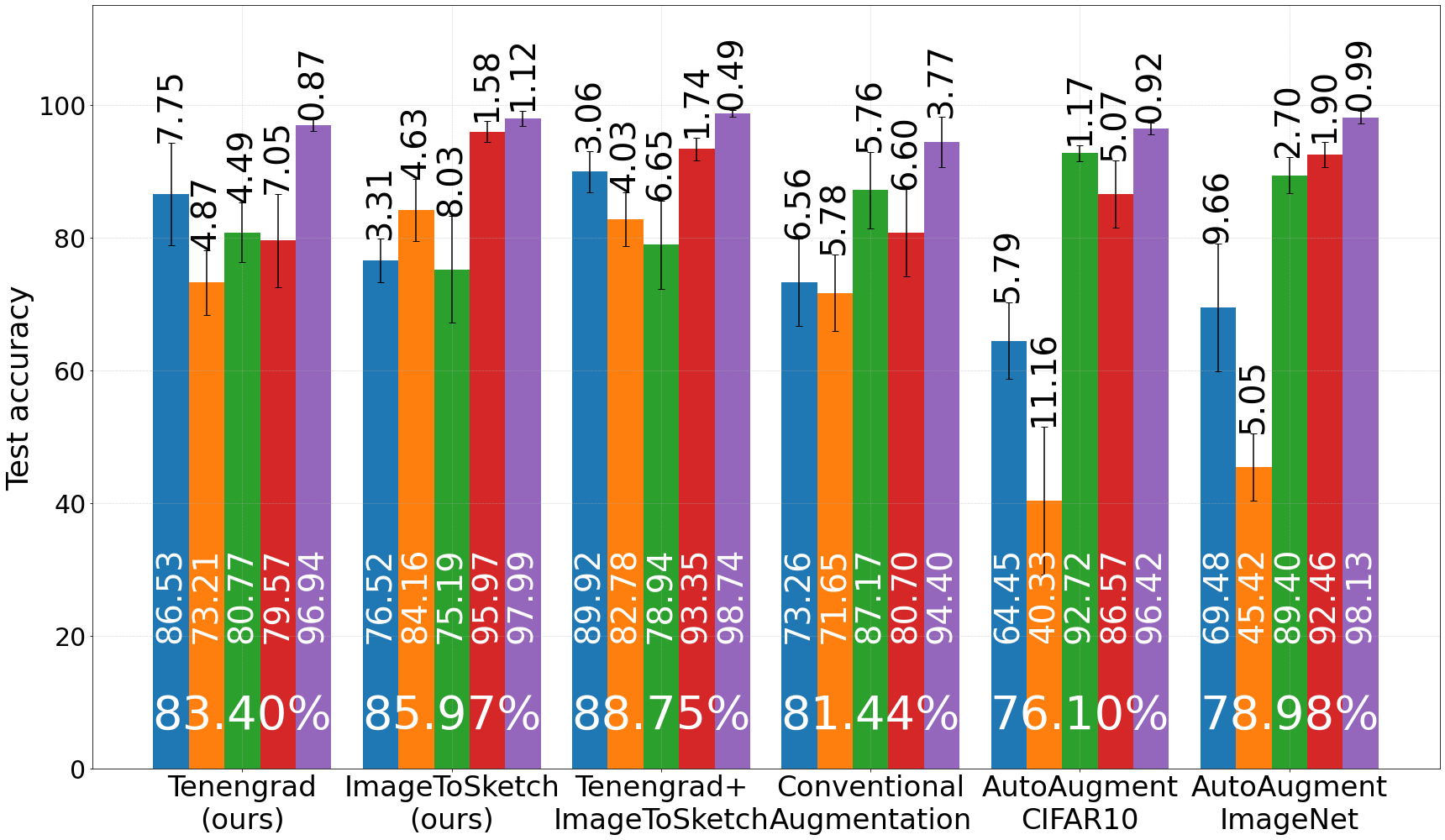}
        \caption{MobileNet-v2}
        \label{fig:eval_mobilenet_v2}
    \end{subfigure}
    \begin{subfigure}{0.325\linewidth}
        \includegraphics[width = \linewidth]{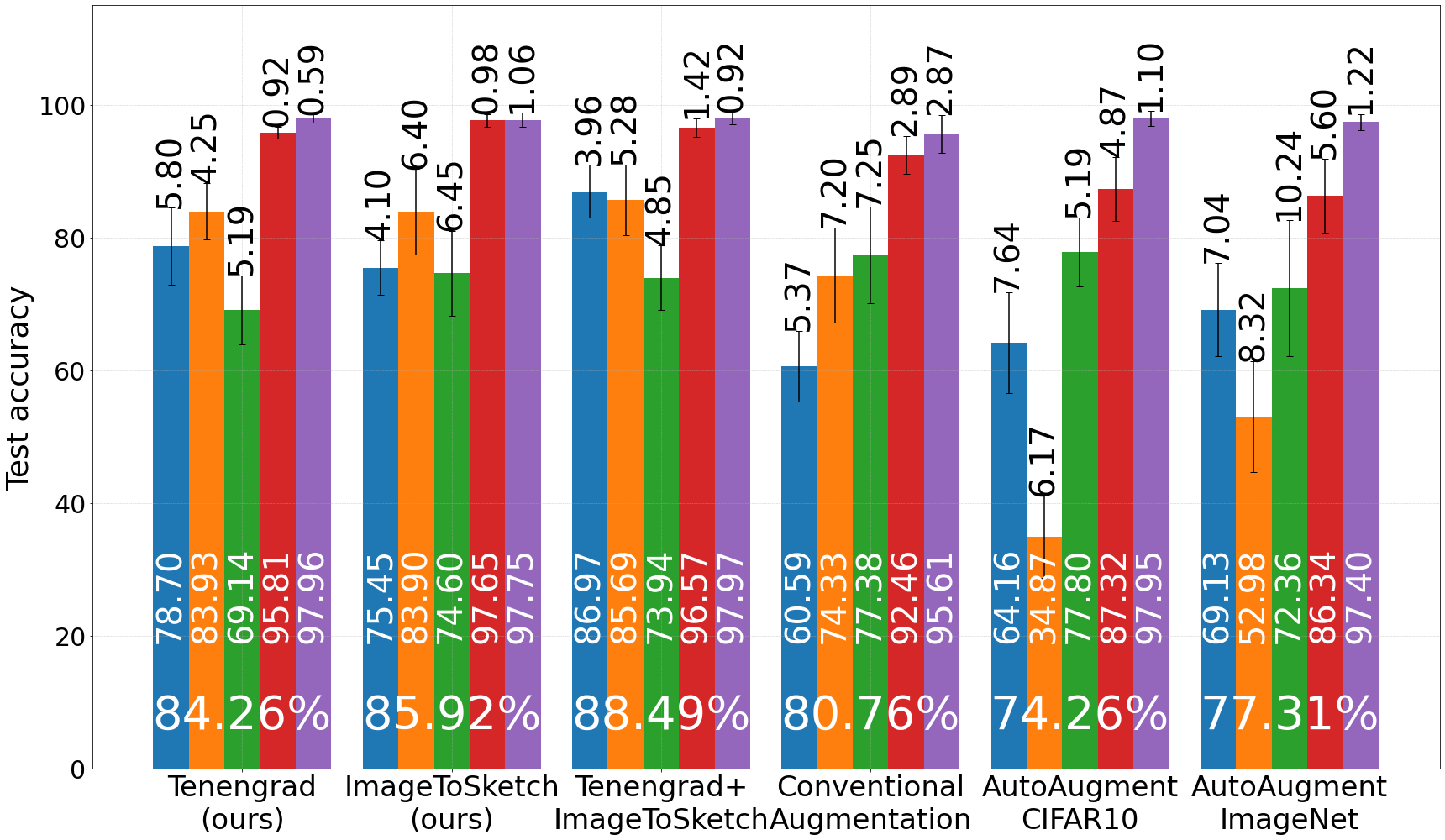}
        \caption{ResNet-18}
        \label{fig:eval_resnet18}
    \end{subfigure}	
    \begin{subfigure}{0.325\linewidth}
        \includegraphics[width = \linewidth]{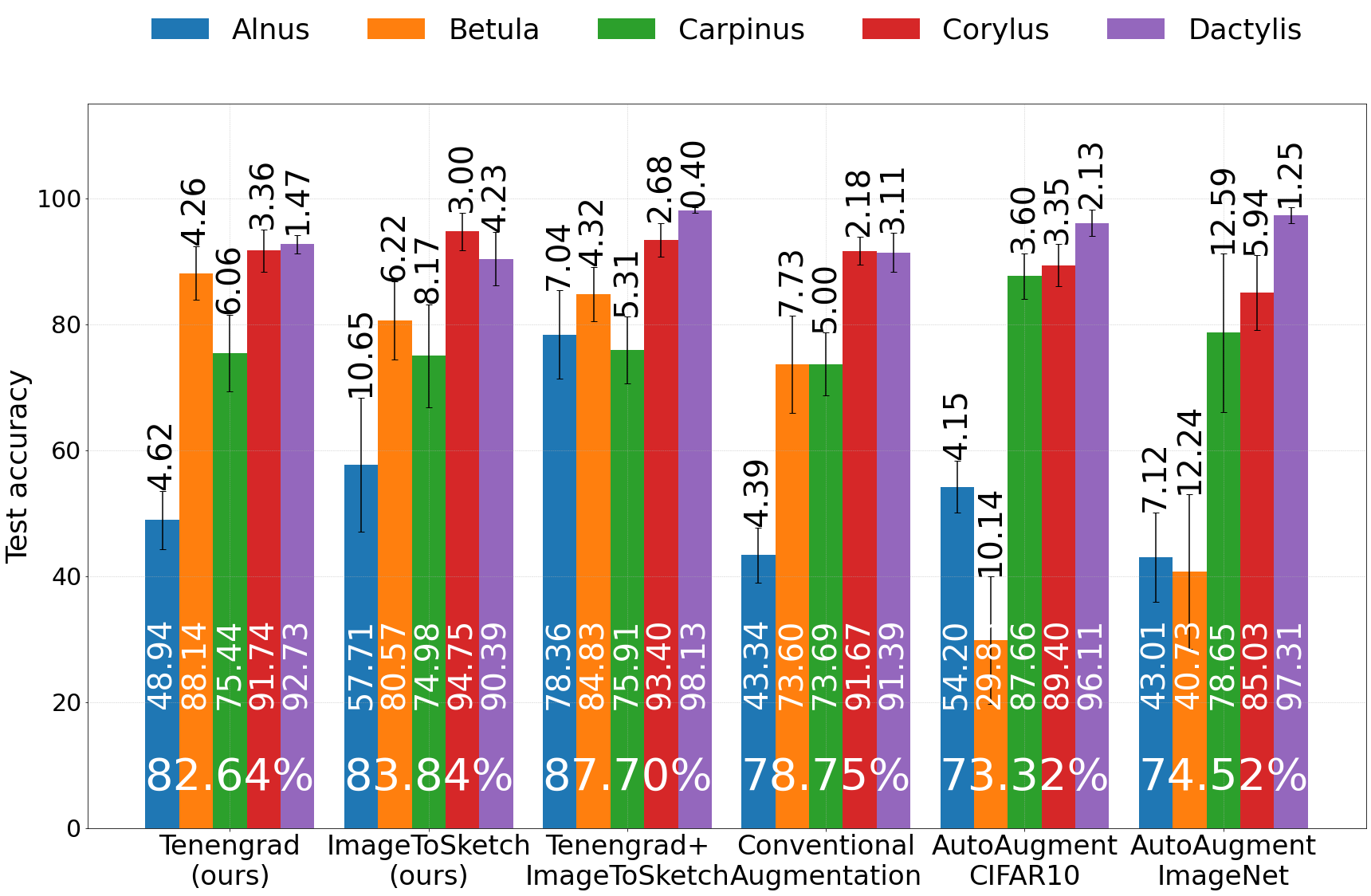}
        \caption{ResNet-50}
        \label{fig:eval_resnet50}
    \end{subfigure}

    \begin{subfigure}{0.325\linewidth}
        \includegraphics[width = \linewidth]{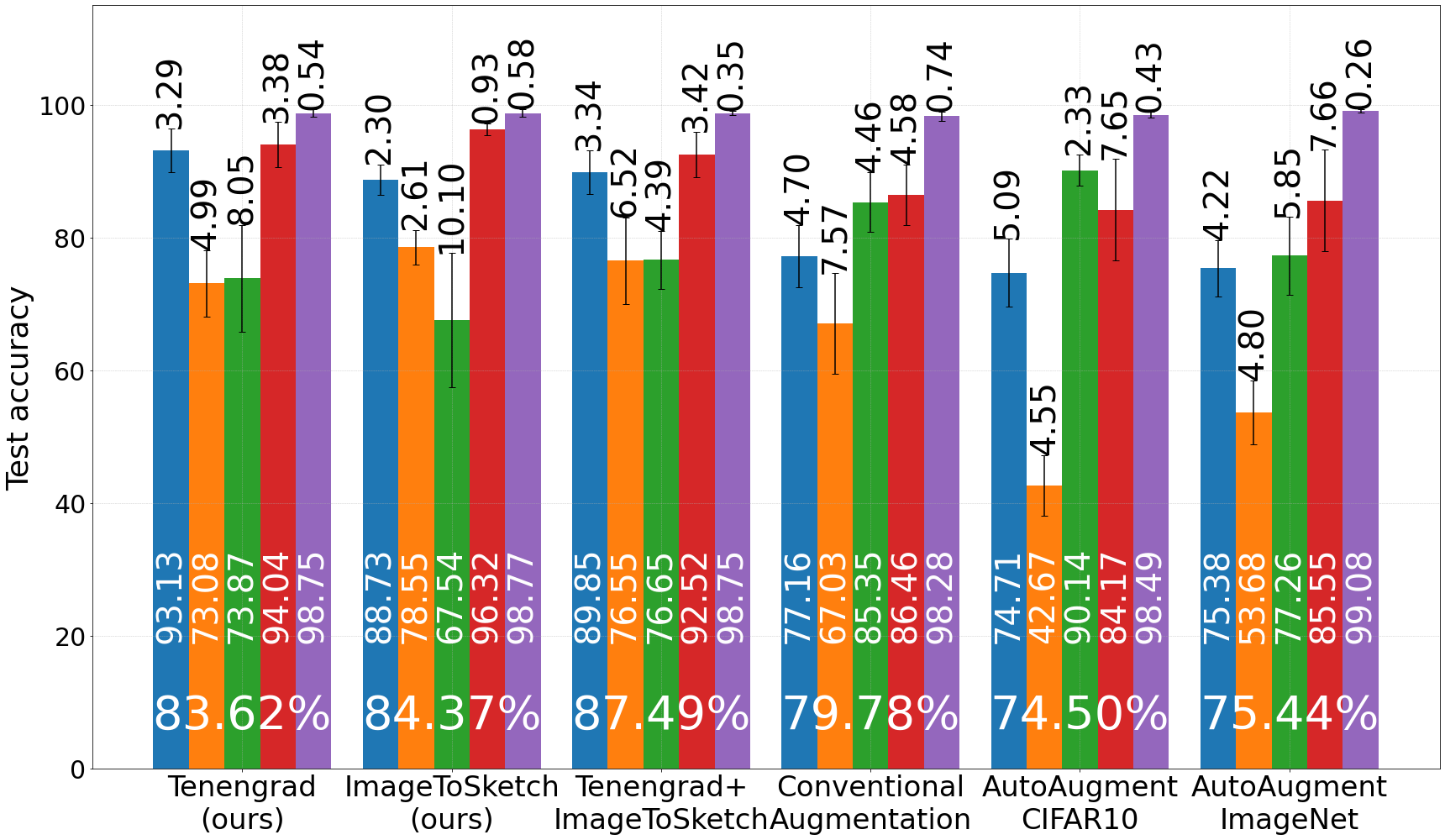}
        \caption{EfficientNet-b0}
        \label{fig:eval_efficientnet_b0}
    \end{subfigure}
    \begin{subfigure}{0.325\linewidth}
        \includegraphics[width = \linewidth]{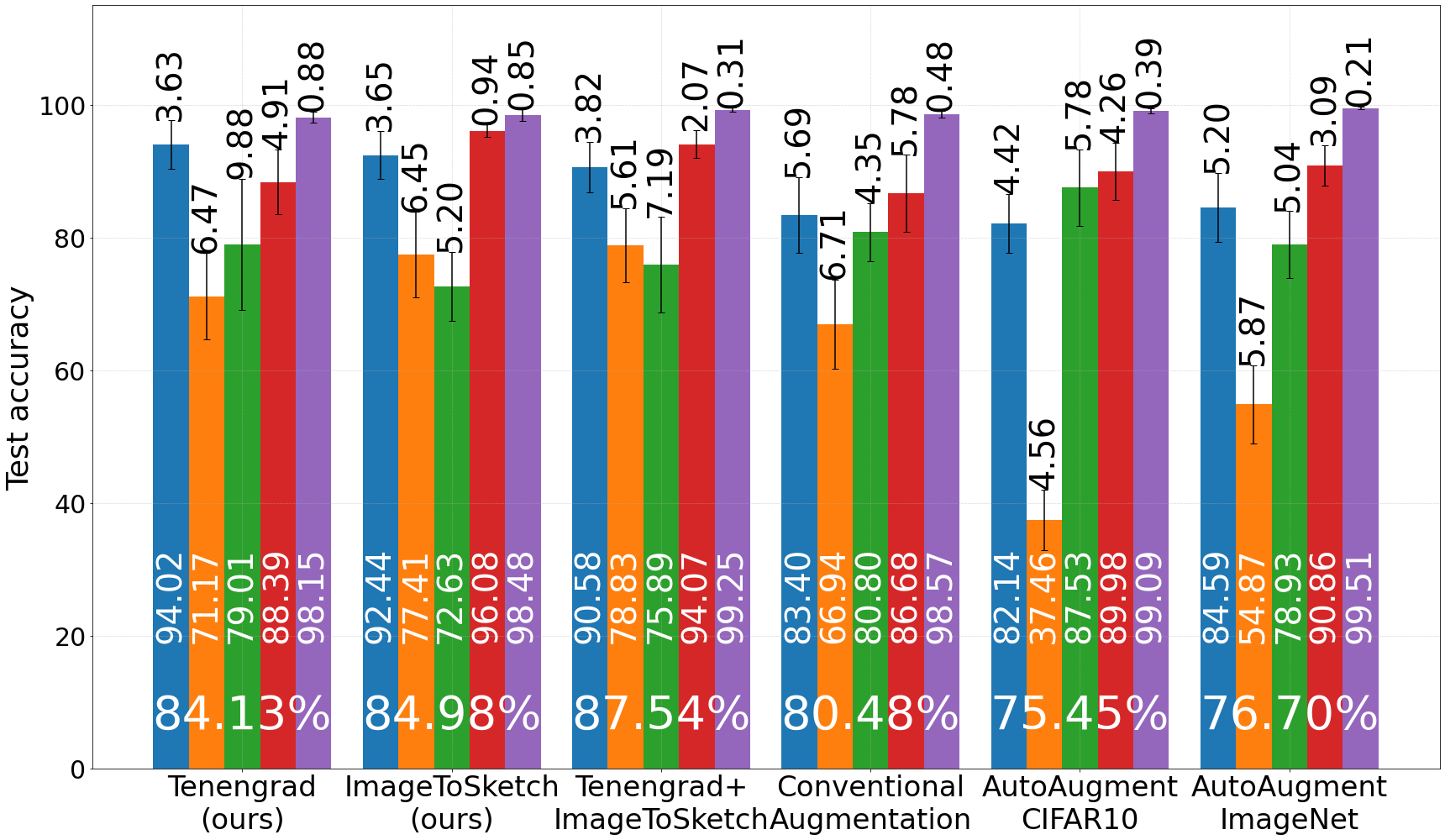}
        \caption{EfficientNet-b4}
        \label{fig:eval_efficientnet_b4}
    \end{subfigure}
    \begin{subfigure}{0.325\linewidth}
        \includegraphics[width = \linewidth]{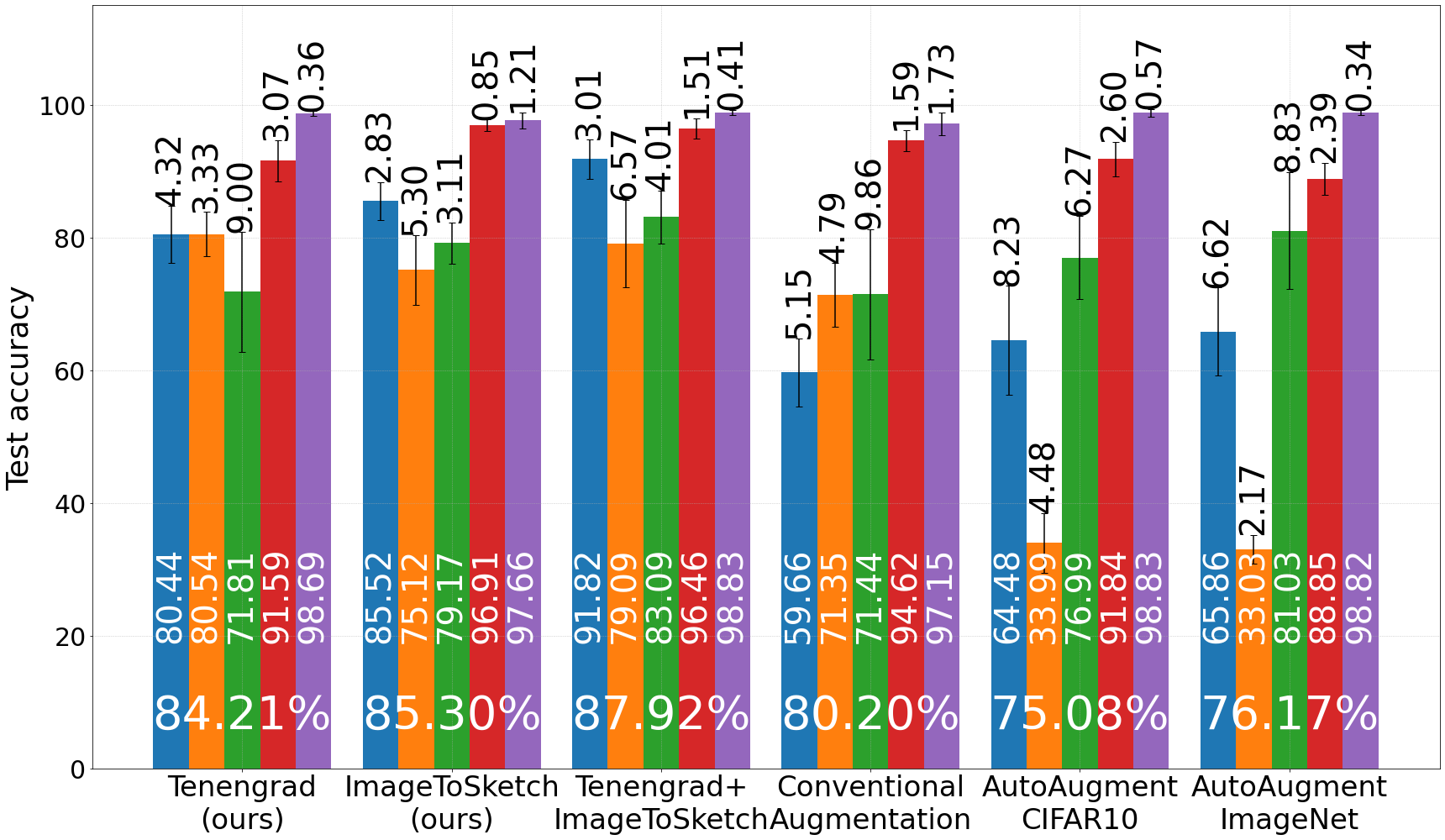}
        \caption{DenseNet-121}
        \label{fig:eval_densenet121}
    \end{subfigure}
    \caption{Comparing augmentation methods when learning to classify pollen using six state-of-the-art CNN architectures. All models were pre-trained on ImageNet and fine-tuned on the pollen library data. We report test accuracy on the field data.}
    \label{fig:variousCNN}
\end{figure*}

All models are tested on the field data, and each experiment is repeated five times to derive meaningful statistics. The obtained mean and standard deviation test set accuracy for each architecture and augmentation method are reported in the bar plots depicted in \Figref{fig:variousCNN}.
The models fine-tuned with geometric augmentations achieve higher accuracy, surpassing other augmentation techniques by a margin between 5\,\% and 14\,\%. The mean accuracy consistently falls within the 82\,\% to 88\,\% range, showcasing an approximate 12\,\% improvement compared to alternative augmentation methods.

The graphs also show that geometric augmentations consistently improve the test set accuracy across all pollen classes. The average accuracy is consistently above 70\,\%. Notably, for the Betula class, the ResNet-50 model exhibits an improvement of up to 60\,\%. The average accuracy for Betula displays greater variability across different architectures when utilizing alternative augmentation methods, highlighting the performance stability offered by the geometric augmentations.

\begin{figure*}[t]
\centering
	\begin{subfigure}{0.31\textwidth}
		\includegraphics[width=\textwidth]{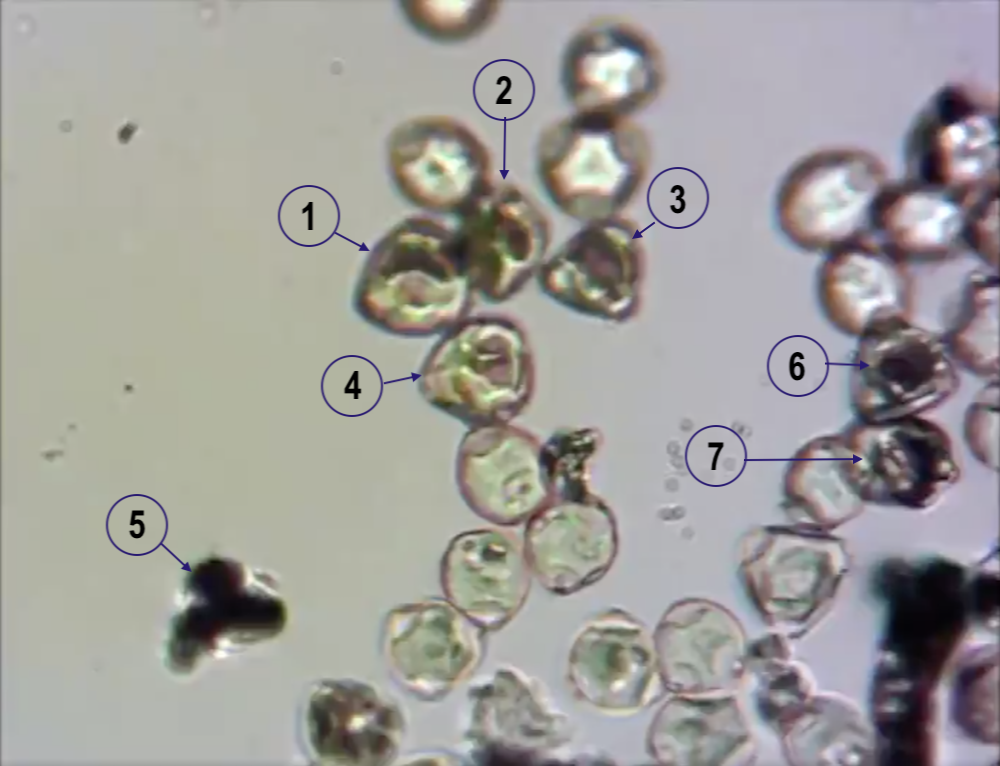}
		\caption{0\,min}
		\label{fig:hydration:0min}
	\end{subfigure}
	\begin{subfigure}{0.31\textwidth}
		\includegraphics[width=\textwidth]{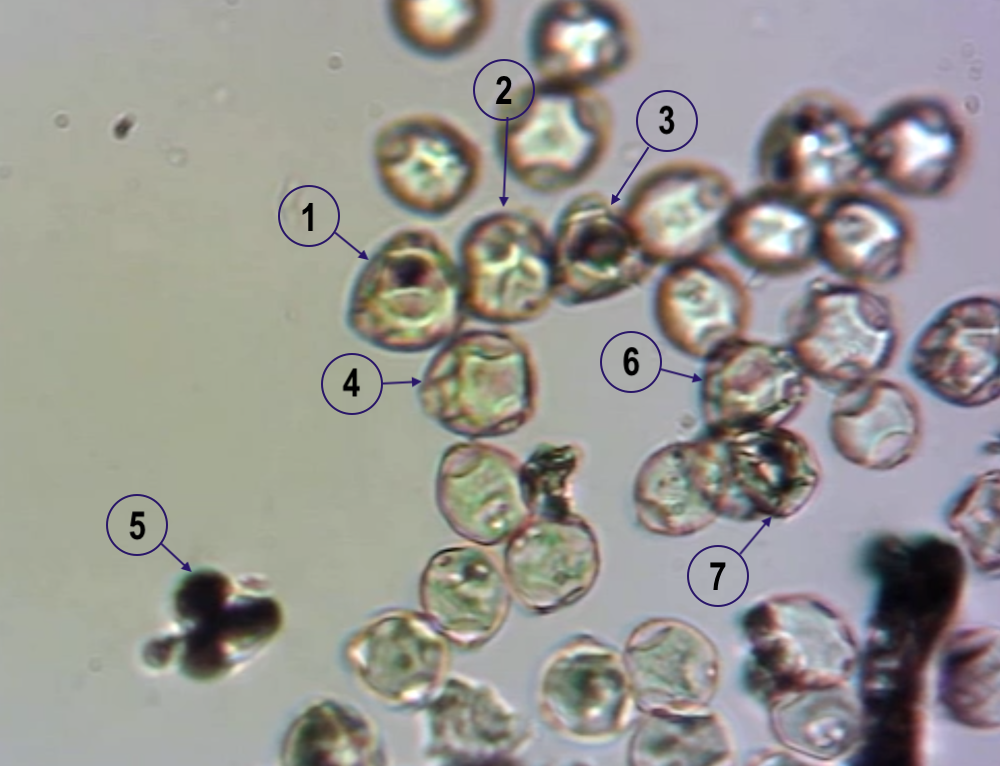}
		\caption{3\,min}
		\label{fig:hydration:3min}
	\end{subfigure}
	\begin{subfigure}{0.31\textwidth}
		\includegraphics[width=\textwidth]{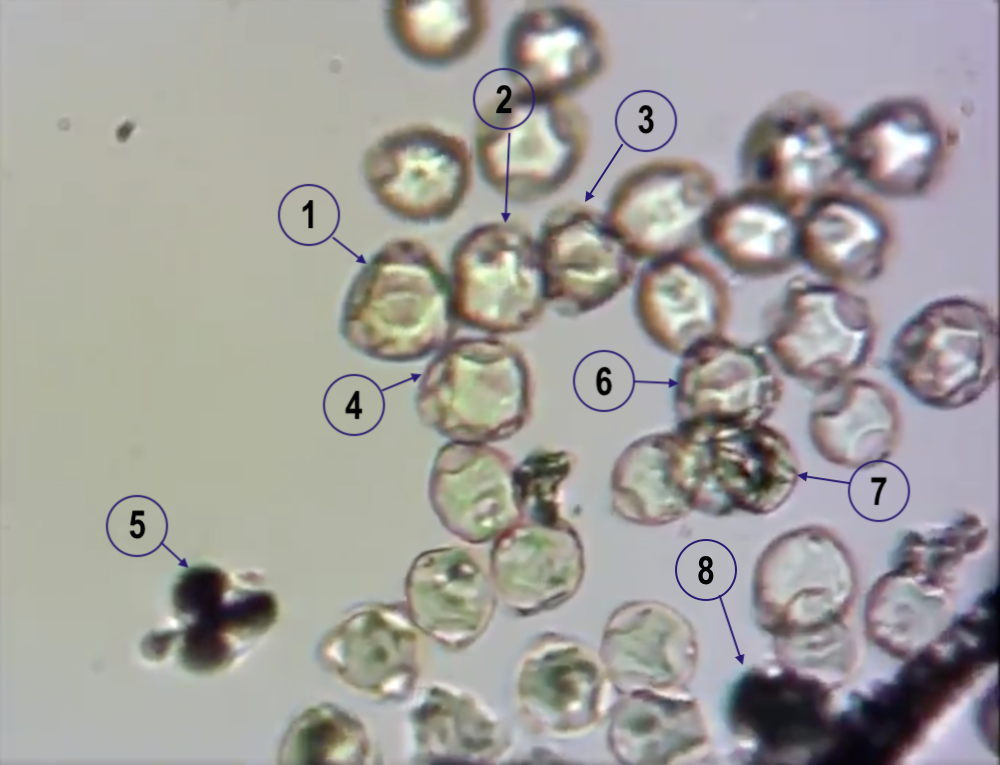}
		\caption{6\,min}
		\label{fig:hydration:6min}
	\end{subfigure}
	\begin{subfigure}{0.31\textwidth}
		\includegraphics[width=\textwidth]{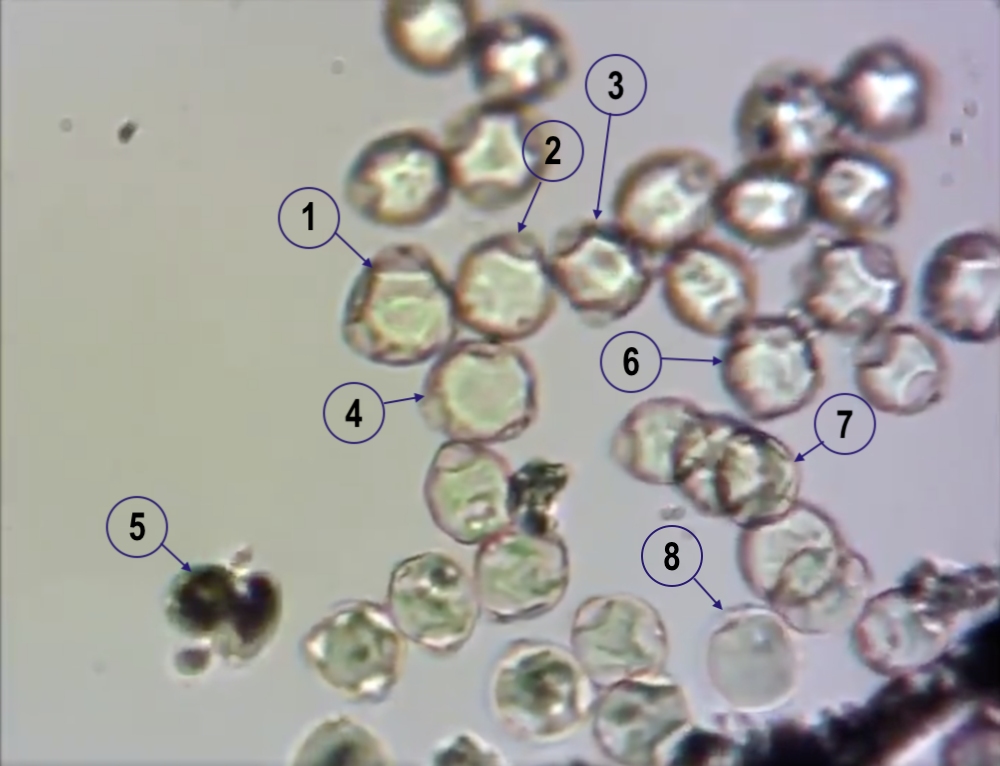}
		\caption{9\,min}
		\label{fig:hydration:9min}
	\end{subfigure}
	\begin{subfigure}{0.31\linewidth}
		\includegraphics[width=\linewidth]{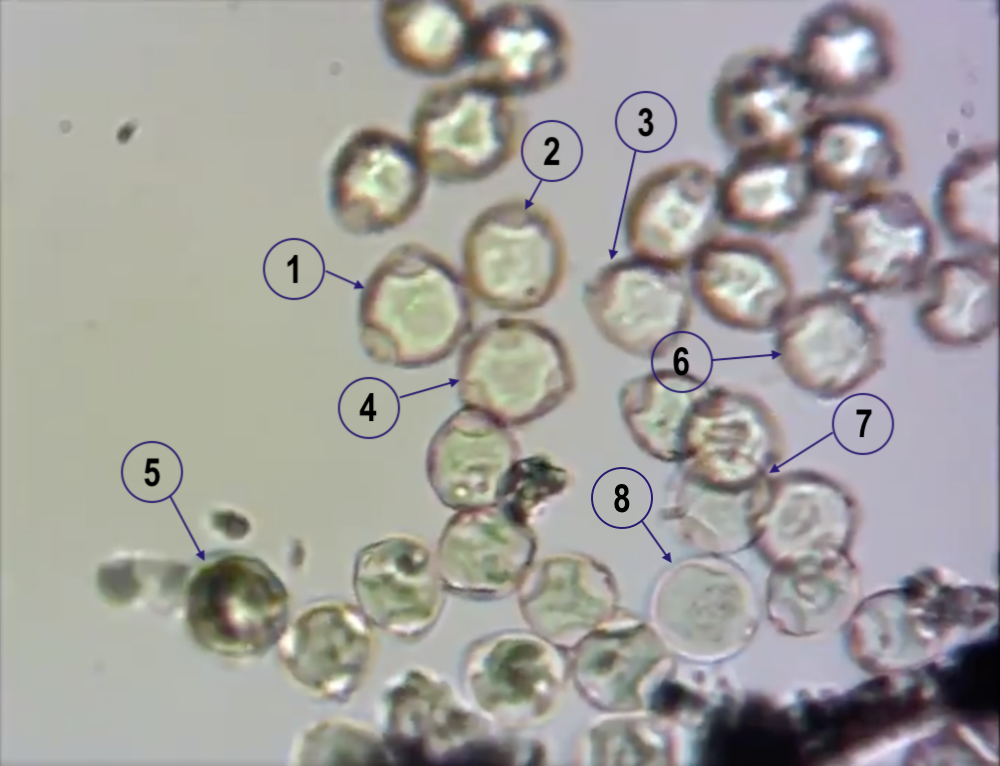}
		\caption{12\,min}
		\label{fig:hydration:12min}
	\end{subfigure}	
	\begin{subfigure}{0.31\textwidth}
		\includegraphics[width=\textwidth]{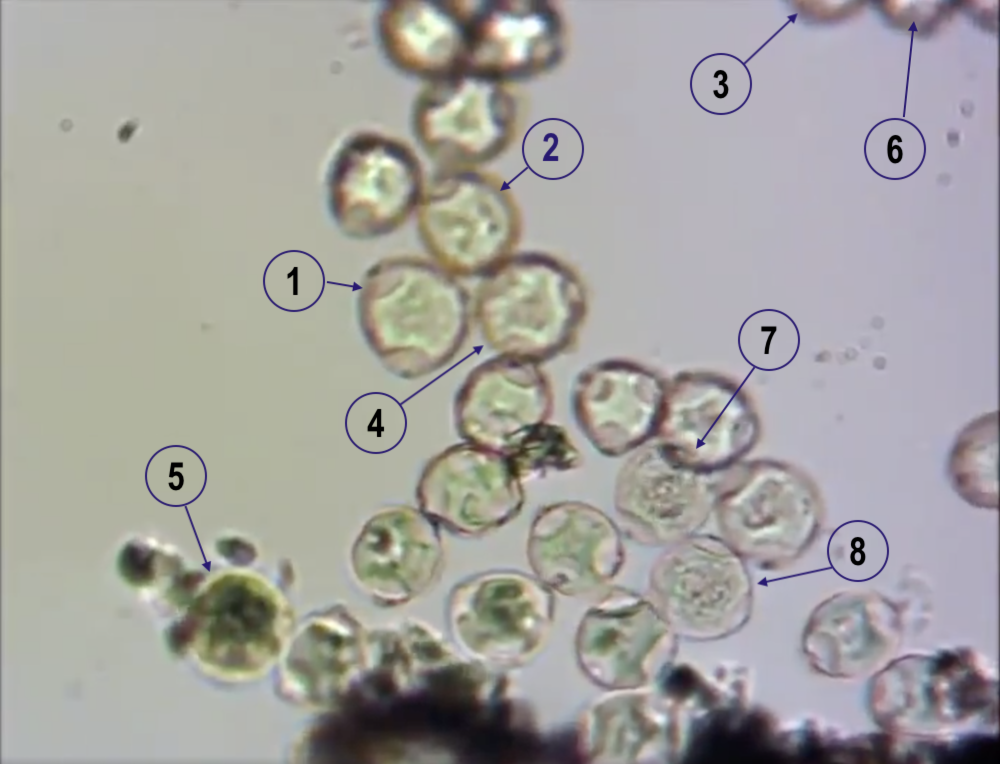}
		\caption{15\,min}
		\label{fig:hydration:15min}
	\end{subfigure}
\caption{Hydrating Corylus (\circled{1}-\circled{6}) and Carpinus (\circled{7}-\circled{8}) pollen grains in glycerine to recover their shapes and internal structures. Grain annotations are consistent across different snapshots.} 
\label{fig:hydration}
\end{figure*}

\subsection{Pollen hydration recovers geometry}
\label{sec:pollen_hydration}
Geometric augmentation methods consistently enhance classification test set accuracy of models of different sizes and architectures. In this subsection we provide evidence that pollen geometry is the cause why our geometric filters work by running an ablation study using the pollen hydration process. Pollen loose their freshness if not immediately being processed: they dry out. Dry pollen have scrambled shape and their internal structure is not well visible under the microscope. 
Hydration in glycerine recovers pollen morphology. \Figref{fig:hydration} shows multiple snapshots capturing different stages of hydration of Corylus and Carpinus pollen grains under the microscope. To ensure a comprehensive evaluation, we chose three specific pollen types: Alnus, Betula, and Taxus, which show strong morphology recovery upon hydration to demonstrate that the network trained with geometric data augmentation can increasingly better learn shape information as hydration advances and grains recover their shape.

We collected pollen samples at three different hydration levels by adding dry pollen into glycerine and recording the hydration process. We split the hydration status into 3 stages: dry ($<$3\,min hydrated in glycerine), half-hydrated (3--9\,min), and hydrated ($>$9\,min). The view of sample pollen grains in these stages is depicted in \Figref{fig:hydration}. In this study, we use the 3-class dataset summarized in \tabref{tab:pollen_status_dataset}.

\begin{table}[!ht]
\centering
\caption{\#samples in the pollen hydration dataset.}
\label{tab:pollen_status_dataset}

\begin{tabular}{lccc}
\toprule
 & \multicolumn{1}{l}{\textbf{Dry pollen}} & \multicolumn{1}{l}{\textbf{Half-hydrated}} & \multicolumn{1}{l}{\textbf{Hydrated}} \\ 
\midrule
\textbf{Alnus}  & 375 & 248 & 127 \\
\textbf{Betula} & 587 & 143 & 126 \\
\textbf{Taxus}  & 308 & 97  & 127 \\ 
\bottomrule
\end{tabular}
\end{table}

\begin{figure*}[t]
\centering

    \begin{subfigure}{0.32\textwidth}
        \includegraphics[width = \textwidth]{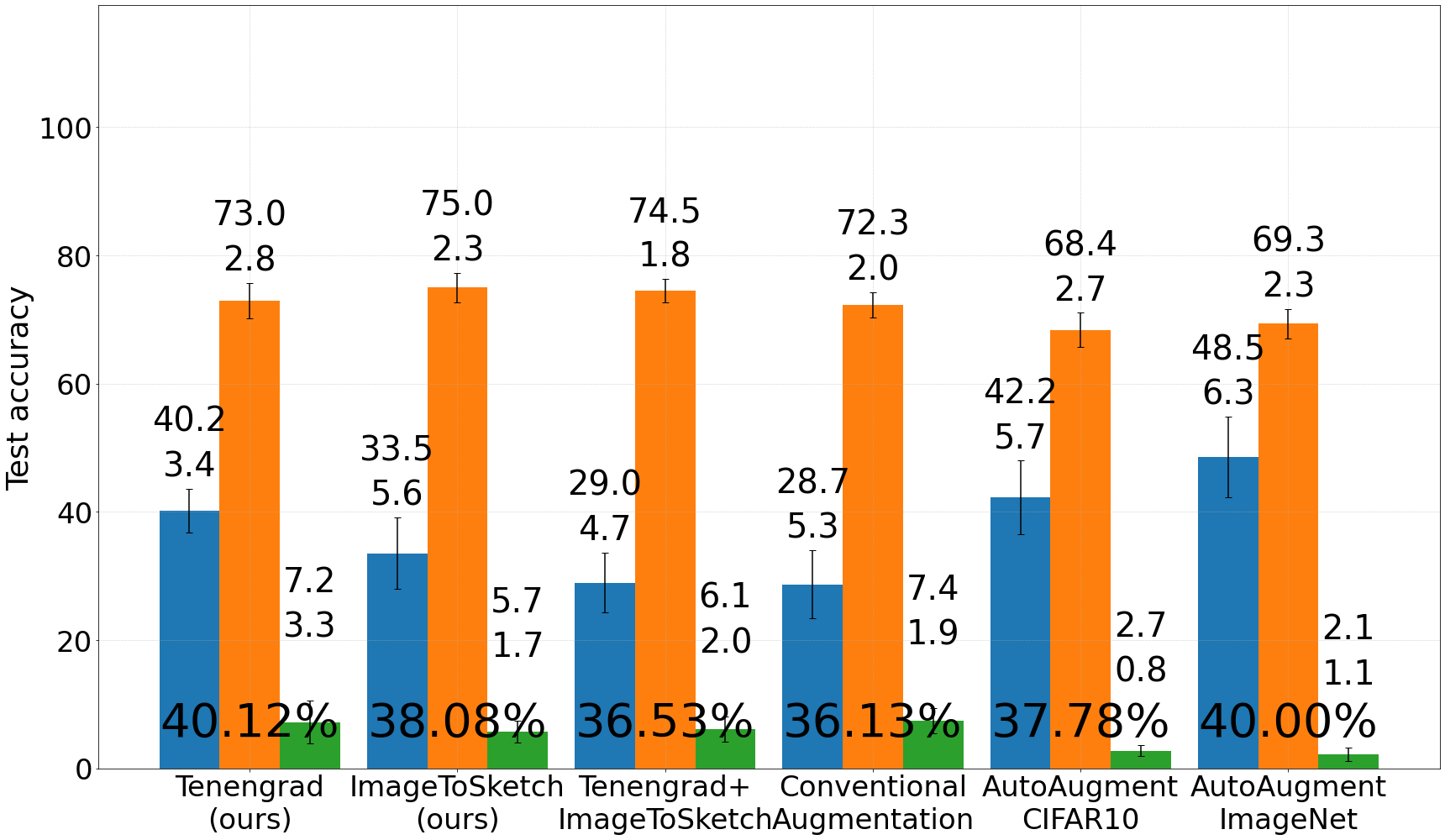}
        \caption{Dry pollen}
        \label{fig:eval_dry_pollen}
    \end{subfigure}
    \begin{subfigure}{0.32\textwidth}
        \includegraphics[width = \textwidth]{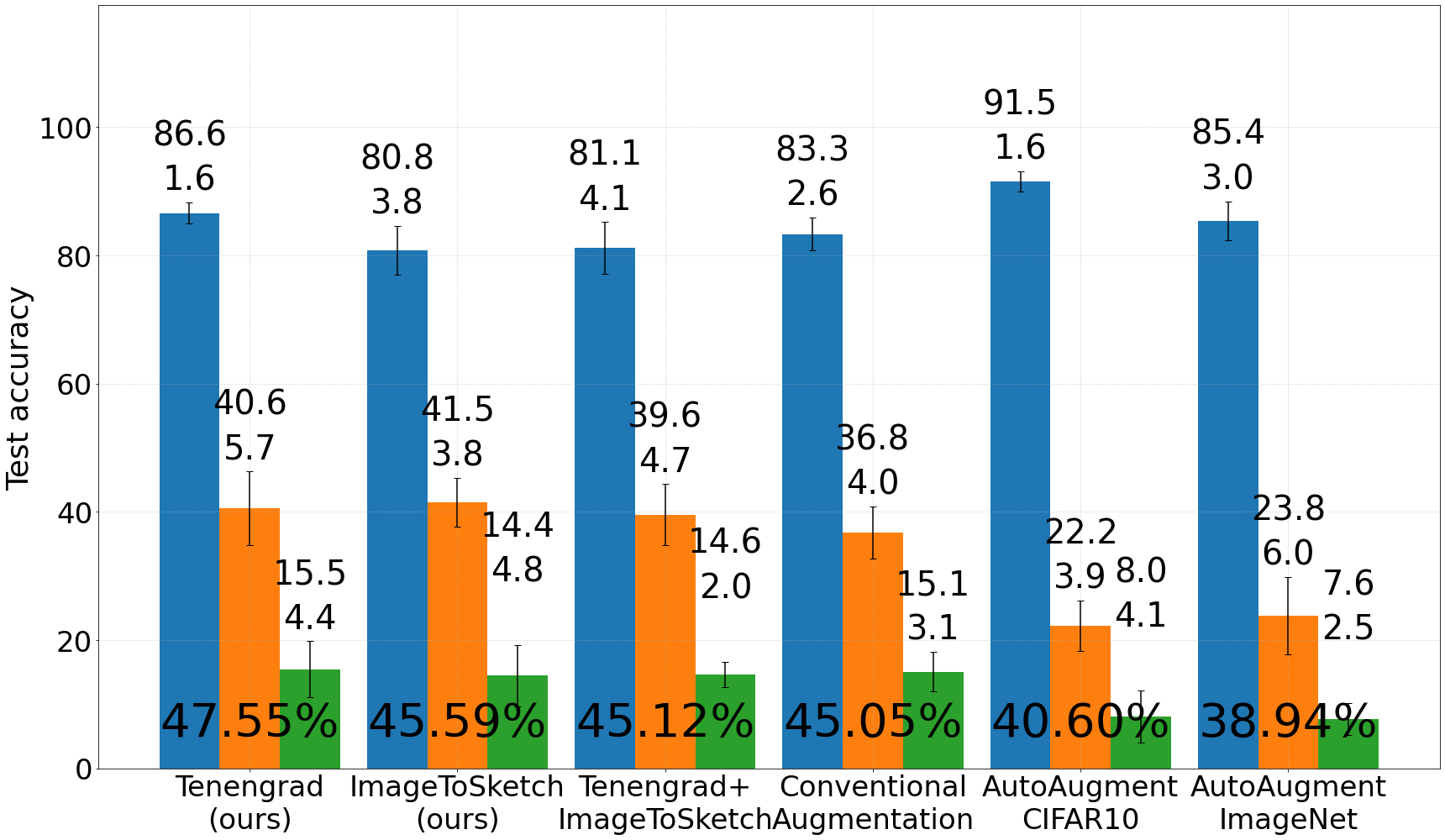}
        \caption{Half-hydrated pollen}
        \label{fig:eval_half_hydrated_pollen}
    \end{subfigure}
    \begin{subfigure}{0.32\textwidth}
        \includegraphics[width = \textwidth]{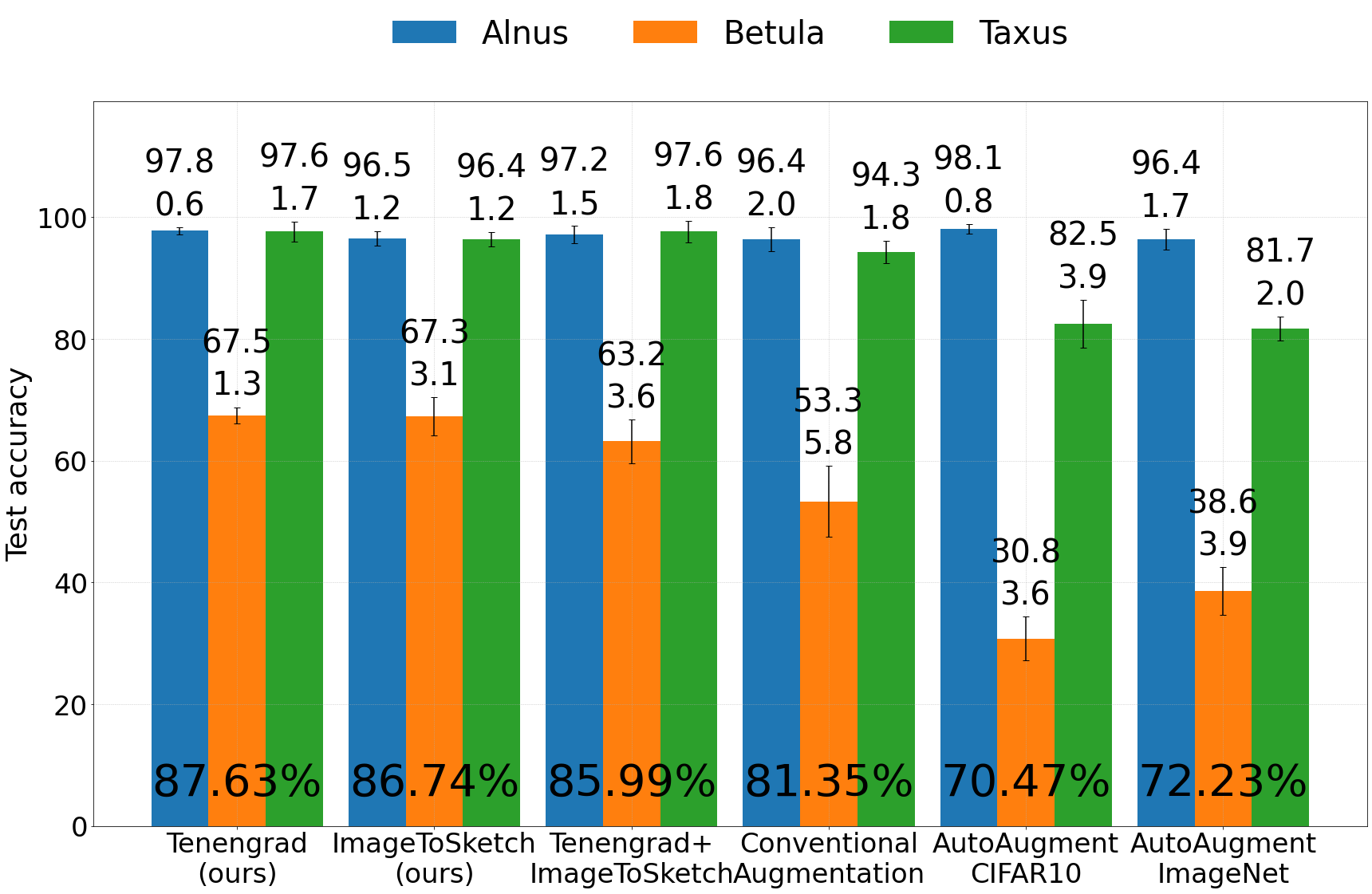}
        \caption{Hydrated pollen}
        \label{fig:eval_hydrated_pollen}
    \end{subfigure}
    \caption{Performance of data augmentation techniques on images of dry, half-hydrated and hydrated pollen. Hydration improves model accuracy on all classes, yet geometric data augmentation is increasingly more effective with hydration as pollen shape and internal structure get recovered by the process.}
    \label{fig:three_stages_evaluation}
\end{figure*}

In this experiment, we opted for ResNet-18 pre-trained on ImageNet as our base model. It was further fine-tuned for 40 epochs on the pollen library data using 3 classes listed above. Each fine-tuning experiment is repeated five times with the same augmentation techniques as described in the first experiment. \Figref{fig:three_stages_evaluation} shows the model test set accuracy on images featuring pollen in three hydration levels. We observe poor performance of all augmentation methods when evaluated on images of dry pollen with an average accuracy of 38\,\%, nearly a random guess. A gradual improvement emerges when pollen is half-hydrated. The geometric augmentation improves the accuracy up to 9\,\% with the Tenengrad augmentation. When the grains are fully hydrated, the accuracy is strongly improved up to 87\,\% for geometric augmentations. 

\subsection{Diversity and affinity measures}
\label{sec:diversity_affinity}
Inspired by \citep{gontijo-lopes2021tradeoffs}, we adopt the two interpretable measures,---affinity and diversity,---to quantify how data augmentation improves the model generalization ability. 

\paragraph{Diversity} refers to the variability and distinctiveness of the augmented data samples compared to the original data. When augmenting data, it is crucial to ensure that the newly generated samples vary enough to provide the model with a broad range of input patterns. This helps the model to generalize to unseen data during training and testing. 
Diversity is defined in ~\citep{gontijo-lopes2021tradeoffs} as a ratio of the training loss when training with augmented data and the training loss when training on original data. Formally,
\begin{equation}
\mathcal{D}[a;m;\mathcal{D}_{train}] = \mathbb{E}_{\mathcal{D}'_{train}}[\mathcal{L}_{train}] \;/\; \mathbb{E}_{\mathcal{D}_{train}}[\mathcal{L}_{train}],
\end{equation}

\noindent where $a$ is an augmentation and $\mathcal{D}'_{train}$ is the augmented training data resulting from applying the augmentation $a$. $\mathcal{L}_{train}$ is the training loss of model $m$ trained on $\mathcal{D}'_{train}$.
Diversity is expected to be high, yet in a way that does not negatively affect model performance.

\paragraph{Affinity} refers to the degree of similarity between the original data and the augmented data. Augmented samples should ideally be similar enough to the original samples such that they do not introduce noise or distort the underlying data distribution. The goal is to expand the dataset without introducing bias or artifacts that could negatively impact the model's ability to learn the underlying patterns.
Affinity, as derived from \citep{gontijo-lopes2021tradeoffs} was adapted to our context. In our scenario, affinity has been adjusted to be the ratio of evaluation accuracy of the model trained on the library data and tested on field data, and the evaluation accuracy of the same model on validation data, \ie the remaining 10\,\% of the data split from the training dataset. More formally, let $a$ be an augmentation and $\mathcal{D}'_{train}$ be the augmented training data resulting from applying the augmentation $a$. Let $m$ be a model trained on $\mathcal{D'}_{train}$ and $\mathcal{A}(m,D)$ denote the model’s accuracy when evaluated on the dataset $\mathcal{D}$. Affinity is given by $$T[a;m;\mathcal{D}_{val}] = \mathcal{A}(m, \mathcal{D}_{field})\;/\;
\mathcal{A}(m, \mathcal{D}_{val}).$$

According to this definition, affinity of one indicates no distribution shift, while a smaller number suggests that the augmented data is out-of-distribution for the model~\citep{gontijo-lopes2021tradeoffs}.

Both affinity and diversity measure the success of data augmentation.
For example, a point with high affinity but low diversity refers to an augmented data sample that closely resembles its corresponding original data but lacks variability compared to other augmented samples. This leads to a potential overfitting on the training data and poor performance on unseen data. Moreover, limited diversity can make the model sensitive to minor changes or noise in the data, reducing its overall robustness and reliability.
Data points with high affinity and high diversity achieve an optimal balance between preserving the original data's integrity and introducing novel variations. This balance helps to prevent overfitting and improve the model's ability to generalize to new, unseen data.

\begin{figure}[t]
  \centering
  \includegraphics[width=.65\linewidth]{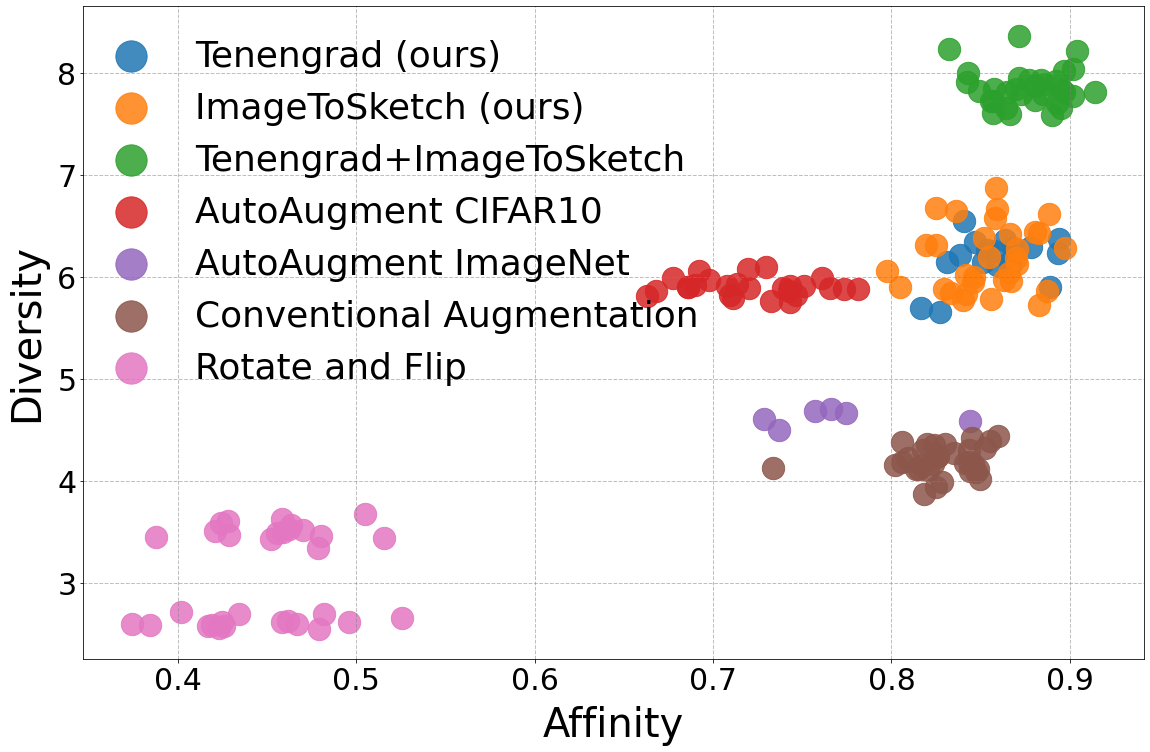}
  \caption{Affinity and diversity measures computed for all data augmentation methods. Proposed geometric augmentation techniques achieve best scores on pollen data (green, orange and blue).}
  \label{fig:affinity_diversity}
\end{figure}



\paragraph{Experimental Results.}
We use the same dataset for training and validation as introduced in the first experiment in \Secref{sec:Datasets}. ResNet-18 pre-trained on ImageNet was the tested architecture. Besides the earlier introduced five augmentation methods described in~\Secref{sec:train_on_different_CNN}, we add Rotate-and-flip augmentation to examine how a simpler augmentation works: 

\begin{Verbatim}[fontsize=\small]
Rotate-and-Flip = Resize + CenterCrop + RandomRotate(90°) + RandomHorizontalFlip + Normalize.
\end{Verbatim}
Each augmentation is used to train the model for 40 epochs. The AutoAugment experiment is repeated at least 5 times. The affinity and diversity are computed for each experiment.



The scatter plot in \Figref{fig:affinity_diversity} shows that, the model trained with our geometric augmentations outperforms the model trained with other augmentation methods. Rotate-and-flip achieves the worst performance among the tested augmentation methods.  Conventional augmentation and  AutoAugment-ImageNet have similar scores. AutoAugment-CIFAR10 does improve the diversity but shifts the distribution of augmented data more than the proposed geometric augmentation techniques.

\section{Conclusion, Limitations, and Outlook}
\label{sec:conclusion}
The paper addresses the distribution shift problem in the domain of pollen identification from microscopic images. 
We introduce two geometric data augmentation techniques to boost the impact of pollen shape and texture information during training. The extensive experimental evaluation conducted on various CNN architectures, including MobileNet-v2, ResNet-18, ResNet-50, EfficientNet-b0, EfficientNet-b4, and DenseNet-121, shows performance improvements due to the use of the proposed augmentation methods across different model sizes and architectures. The ablation study shows the models' ability to increasingly outperform other methods as pollen grains get hydrated. Finally, we adapt the affinity and diversity measures~\citep{gontijo-lopes2021tradeoffs} to evaluate the effectiveness of the proposed geometric augmentation techniques. This methodology enables a more comprehensive and insightful analysis of the augmentation's impact on the model performance and generalization in pollen classification based on microscopic images.

The proposed augmentation techniques to mititage distribution shift are untimately domain-specific, and work well on low-contrast images, and similar object data. 
%
We show that expert knowledge and targeted data augmentations allow developing more robust models. However, the general methodology which could be used to tackle other domains is not yet clear. In the future, we plan to better understand the differences in data distribution and contribute to the design of a  general framework to build robust data-driven models, also for other domains.

\section{Acknowledgments}
This research is partly funded by the Vietnam International Education, Cooperation Department, Ministry of Education and Training, Project 911. The results were computed using computational resources of HLR resources of the Zentralen Informatikdienstes of TU Graz. The authors would like to thank Rahim Entezari for the discussion of affinity and diversity measures.


\bibliography{arxiv}
\bibliographystyle{apalike-ejor}

\appendix
\section*{Appendix}
\section{Implementation Details}
\label{app:traning_parameters}

\tabref{hyperparam} summarizes the set of hyper-parameters used to train different networks throughout this work. All experiments were repeated 5 times.

\begin{table}[h]
\caption{Training hyper-parameters for all model architectures.}
\label{hyperparam}
\centering
\begin{tabular}{ll}
\hline
\textbf{\textbf{Parameter}} & \textbf{\textbf{Value}}                                                \\ \hline
Batch Size       & 32               \\
Input Image Size & 224              \\
Loss Function    & CrossEntropyLoss \\
Optimizer        & Adam             \\
Learning rate    & 10$^{-3}$         \\
Learning Rate Scheduler     & \texttt{lr\_scheduler.StepLR(optimizer\_ft, step\_size=30, gamma=0.1)} \\ \hline
\end{tabular}
\end{table}

\end{document}